\documentclass{article}
\usepackage{microtype}
\usepackage{graphicx}
\usepackage{subcaption}
\usepackage{booktabs} 
\usepackage{booktabs}
\usepackage{multirow}
\usepackage{wrapfig}
\usepackage[dvipsnames]{xcolor}

\usepackage{tcolorbox}
\tcbuselibrary{skins}           
\tcbuselibrary{breakable}    

\usepackage{hyperref}

\usepackage{xcolor}

\usepackage[preprint]{neurips_2026}
\usepackage{xspace}
\newcommand{\methodname}{\textit{SwitchSD}\xspace}

\usepackage[utf8]{inputenc} 
\usepackage[T1]{fontenc}    
\usepackage{hyperref}       
\usepackage{url}            
\usepackage{booktabs}       
\usepackage{amsfonts}       
\usepackage{nicefrac}       
\usepackage{microtype}      
\usepackage{xcolor}         
\usepackage{amsmath}
\usepackage{amssymb}
\usepackage{mathtools}
\usepackage{amsthm}

\title{To Copy or Not to Copy: Controlling Speculative Decoding via Intrinsic Model Signals}

\author{%
  \bf Roy Eisenstadt\textsuperscript{1,2},\;
  \bf Ido Cohen\textsuperscript{2},\;
  \bf Edo Cohen-Karlik\textsuperscript{2} \\
  \bf Lior Wolf\textsuperscript{1},\;
  \bf Itamar Zimerman\textsuperscript{1,2}
}

\begin{document}

\maketitle

{%
  \renewcommand{\thefootnote}{}%
  \footnotetext{\textsuperscript{1}Tel Aviv University \qquad \textsuperscript{2}Stealth Startup, Tel Aviv}%
}%

\begin{abstract}

  Speculative Decoding (SD) has significantly accelerated Large Language Model (LLM) inference, yet existing approaches face a fundamental tradeoff between two drafting strategies: neural drafting and context-based copying. Neural drafts (e.g., EAGLE3) provide robust performance across diverse text settings, while copy-based methods achieve higher speedups in copy-intensive regimes by generating candidates faster and exploiting long repetition spans for near-perfect speculation. We analyze existing copy-based methods and find that they are prone to \textit{accidental repetitions} where surface-level $n$-gram overlap does not reflect a structural intent to copy, leading to false-positive triggers that ultimately degrade throughput. We introduce \methodname, an adaptive framework that treats copying as a \textit{latent control signal} of the LLM. By training lightweight probes on the target model's internal representations, SwitchSD identifies genuine \textit{copy-intent} with high precision (AUC $>$ 0.99). This allows the system to dynamically switch between neural drafting (e.g., EAGLE) and context-based copying. Our results across Llama and Qwen families demonstrate throughput gains of up to 15\% over state-of-the-art baselines like EAGLE3, effectively turning ``copying'' from a noisy heuristic into a principled, model-aware decoding regime.

\end{abstract}

\vspace{-5pt}
\section{Introduction}
\vspace{-2pt}

Large Language Models (LLMs) have redefined state-of-the-art performance across nearly all NLP domains, often exhibiting emergent reasoning abilities \citep{kojima2022large}. However, these gains are increasingly gated by the computational cost of autoregressive decoding. This bottleneck is most acute in the \textit{test-time scaling regime} \citep{snell2024scaling,welleck2024decoding}, where generating exhaustive chain-of-thought (CoT) sequences significantly inflates latency. As LLMs move toward deeper reasoning, the sequential nature of decoding remains the primary hurdle to real-world deployment.



Speculative decoding (SD) has emerged as the standard remedy for this bottleneck \citep{leviathan2023fast,chen2023accelerating}, utilizing a lightweight draft model to propose token sequences that the target model verifies in parallel. Beyond standard neural drafts, non-neural approaches such as \textit{Prompt Lookup Decoding (PLD)} \citep{saxena2023prompt} and retrieval-based speculators \citep{quan-etal-2025-rasd} have attempted to bypass neural overhead by copying directly from the context via $n$-gram search. Building on these ideas, recent innovations like \textit{CopySpec} \citep{dumitru2025copyspec} introduced a hybrid heuristic: if a prefix has appeared previously in the context, the system speculates the subsequent continuation, falling back to a neural draft strategy if no previous occurrence is found. While intuitive, this approach relies on a brittle assumption that surface-level $n$-gram overlap always implies a structural intent to reproduce context.

\begin{figure*}[t]
  \centering
  \includegraphics[width=0.96\linewidth]{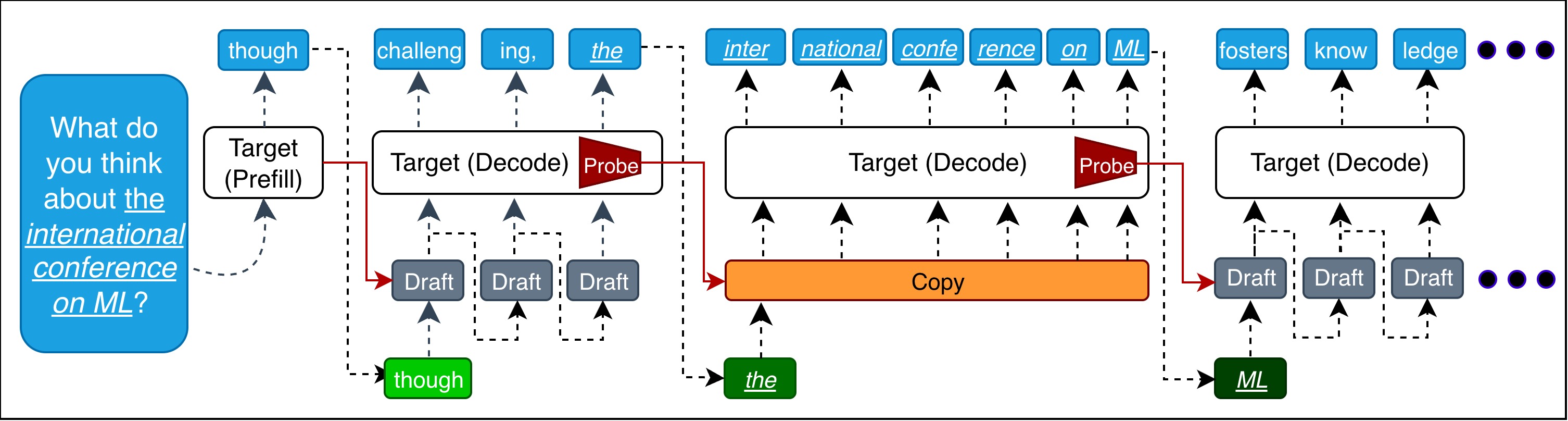}
   \caption{\textbf{Token generation process under our method.} After every verification step by the target model, given the final accepted token (shown in green), there are three possible cases: (i) No repetition contains the currently generated token. In this case, a standard neural draft model is used. Otherwise, a repetition containing the current token exists. In this case: (ii) if the probe (shown in red) identifies copy-intent, speculative tokens are drafted by copying from the context, and (iii) if no copy-intent is identified, the neural draft model is used. The neural draft model is shown in gray, the copy draft in orange, generated tokens in blue, and repeated input tokens with underline. The example in the figure contains the three cases in their numbered order.}
  
  \label{fig:specSwtich}
\vspace{-4pt}
\end{figure*}

\begin{figure}[t]
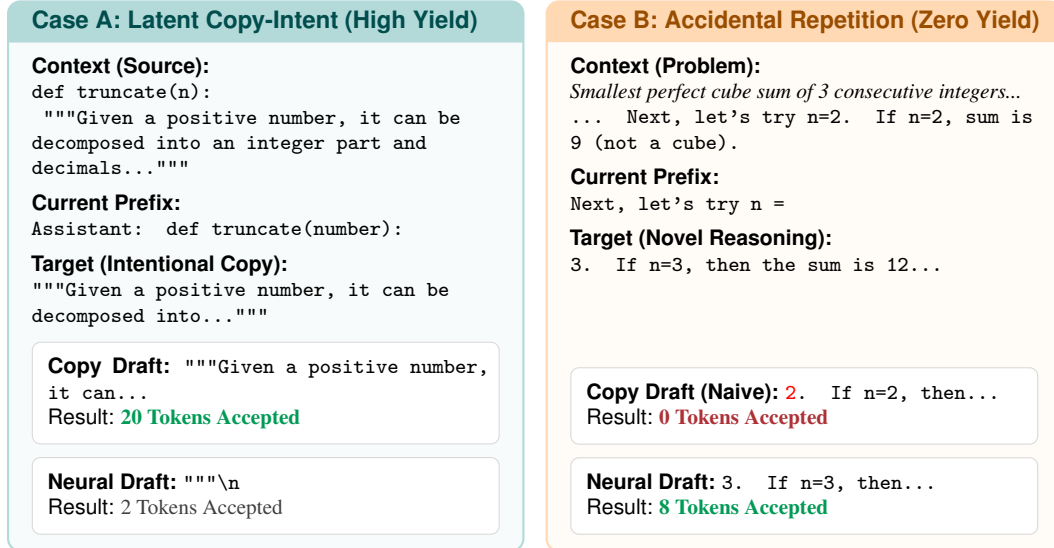

  \centering
  \begin{subfigure}[t]{0.49\textwidth}
    \centering
    \footnotesize
    \begin{tcolorbox}[
      enhanced,
      colback=teal!3,
      colframe=teal!30,
      title=\textsf{\textbf{Case A: Latent Copy-Intent (High Yield)}},
      fonttitle=\bfseries\small,
      coltitle=teal!60!black,
      arc=1.5mm, boxrule=0.8pt,
      left=2mm, right=2mm, top=1mm, bottom=1mm,
      equal height group=qualfig, 
      space to=\myspaceA          
    ]
    \raggedright
    \textsf{\textbf{Context (Source):}} \\
    \texttt{def truncate(n):} \\
    \texttt{\textbf{ """Given a positive number, it can be decomposed into an integer part and decimals..."""}} \\[4pt]
    \textsf{\textbf{Current Prefix:}} \\
    \texttt{Assistant: def truncate(number):} \\[4pt]
    \textsf{\textbf{Target (Intentional Copy):}} \\
    \texttt{\textbf{"""Given a positive number, it can be decomposed into..."""}} 
    
    \vspace{\myspaceA} 

    \begin{tcolorbox}[colback=white, boxrule=0.4pt, colframe=gray!30, left=1mm, right=1mm, top=1mm, bottom=1mm]
      \textbf{\textsf{Copy Draft:}} \texttt{"""Given a positive number, it can...} \\
      \textsf{Result:} \textcolor{ForestGreen}{\textbf{20 Tokens Accepted}}
    \end{tcolorbox}
    
    \begin{tcolorbox}[colback=white, boxrule=0.4pt, colframe=gray!30, left=1mm, right=1mm, top=1mm, bottom=1mm]
      \textbf{\textsf{Neural Draft:}} \texttt{"""\textbackslash n} \\
      \textsf{Result:} \textcolor{darkgray}{2 Tokens Accepted}
    \end{tcolorbox}
    \end{tcolorbox}
  \end{subfigure}
  \hfill
  \begin{subfigure}[t]{0.49\textwidth}
    \centering
    \footnotesize
    \begin{tcolorbox}[
      enhanced,
      colback=orange!3,
      colframe=orange!30,
      title=\textsf{\textbf{Case B: Accidental Repetition (Zero Yield)}},
      fonttitle=\bfseries\small,
      coltitle=orange!60!black,
      arc=1.5mm, boxrule=0.8pt,
      left=2mm, right=2mm, top=1mm, bottom=1mm,
      equal height group=qualfig, 
      space to=\myspaceB          
    ]
    \raggedright
    \textsf{\textbf{Context (Problem):}} \\
    \textit{Smallest perfect cube sum of 3 consecutive integers...} \\
    \texttt{... \textbf{Next, let's try n=2.} If n=2, sum is 9 (not a cube).} \\[4pt]
    \textsf{\textbf{Current Prefix:}} \\
    \texttt{\textbf{Next, let's try n =}} \\[4pt]
    \textsf{\textbf{Target (Novel Reasoning):}} \\
    \texttt{3. If n=3, then the sum is 12...}
    
    \vspace{\myspaceB} 
    
    \begin{tcolorbox}[colback=white, boxrule=0.4pt, colframe=gray!30, left=1mm, right=1mm, top=1mm, bottom=1mm]
      \textbf{\textsf{Copy Draft (Naive):}} \texttt{\textcolor{red}{2}. If n=2, then...} \\
      \textsf{Result:} \textcolor{Maroon}{\textbf{0 Tokens Accepted}}
    \end{tcolorbox}
    
    \begin{tcolorbox}[colback=white, boxrule=0.4pt, colframe=gray!30, left=1mm, right=1mm, top=1mm, bottom=1mm]
      \textbf{\textsf{Neural Draft:}} \texttt{3. If n=3, then...} \\
      \textsf{Result:} \textcolor{ForestGreen}{\textbf{8 Tokens Accepted}}
    \end{tcolorbox}
    \end{tcolorbox}
  \end{subfigure}

  \caption{\textbf{Latent Copy-Intent vs. Accidental Repetition.} 
  \textbf{(Left)} \methodname identifies copy-intent in docstrings.
  \textbf{(Right)} In reasoning tasks, $n$-gram overlaps are incidental; \methodname correctly defaults to neural drafting.}
  \label{fig:Qualitative_accidental_vs_copy_intent}
  \vspace{-6pt}
\end{figure}

Our analysis reveals a critical flaw in these surface-level heuristics: they cannot distinguish between \textit{true copy-intent} and \textit{accidental repetitions}. As shown in Figure~\ref{fig:Qualitative_accidental_vs_copy_intent}, repeated prefixes often arise from coincidental associative patterns (e.g., in math reasoning) rather than a deliberate reproduction of structured text. Triggering a copy-based draft in these accidental regimes results in immediate rejection, incurring a ``speculation tax'' that degrades throughput compared to standard neural drafting. We argue that copying is not a surface phenomenon but is driven by a \textit{latent control signal} encoded within the model’s internal representations.

Motivated by this, we propose \textbf{\methodname}, the first model-aware SD framework that replaces noisy heuristics with a principled, representation-driven decision mechanism. By extracting latent copy-signals directly from the target model’s \textit{intrinsic representations}, \methodname identifies the intent to copy which allows for dynamic orchestration: speculating via context-copying only when the model is in a ``copy-regime,'' and falling back to state-of-the-art neural speculators (e.g., EAGLE3~\citep{li2025eagle}) otherwise. This selective mechanism eliminates false-positive copy attempts, transforming context-copying from a brittle heuristic into a robust, high-yield acceleration strategy.

\textbf{Our main contributions}  are as follows: (i) We introduce \methodname, an adaptive SD framework that dynamically switches between neural drafting and context copying using lightweight probes over the target model’s hidden representations. Our approach is orthogonal to existing SD methods and consistently improves strong baselines, including EAGLE3, by up to 15\% throughput across multiple benchmarks and model families. (ii) We show that SD contains heterogeneous decoding regimes with substantially different acceptance-length characteristics. In particular, copy-favorable regions produce significantly longer accepted continuations than standard generative regions. Exploiting this structure enables more effective draft-policy selection and improved tuning of speculative lookahead. (iii) Through extensive ablations and analysis, we demonstrate that internal model representations provide reliable control signals for adaptive SD and systematically analyze the impact of probe design, training data, threshold selection, and drafting strategies on overall decoding efficiency.

\section{Background and Related Work}
This section establishes the formal framework for SD and context-copying mechanisms. 

\paragraph{Auto-Regressive Transformers.}
Modern LLMs \citep{vaswani2017attention} consist of stacked Transformer blocks where hidden representations $H^\ell$ are transformed through self-attention and MLP sublayers. Under the pre-normalization formulation, a block is defined as:
{\small
\begin{equation}
\label{eq:transformer_block}
\begin{aligned}
\tilde{H}^{\ell} &= H^{\ell} + \mathrm{Attn}\!\left(\mathrm{LN}(H^{\ell})\right), \\
H^{\ell+1} &= \tilde{H}^{\ell} + \mathrm{FFN}\!\left(\mathrm{LN}(\tilde{H}^{\ell})\right).
\end{aligned}
\end{equation}
}
In autoregressive LLMs, tokens $x_t$ are generated sequentially, conditioned on the full preceding context. Our work specifically utilizes the intermediate representations $H^\ell$ at layer $\ell$ to detect latent decoding regimes.

\paragraph{Speculative Decoding.}
SD \citep{leviathan2023fast,chen2023accelerating} accelerates inference by using a lightweight draft model to propose $k$ candidate tokens, which are verified in parallel by the target model. Efficiency is governed by the acceptance rate $\alpha = 1 - KL(p, q)$, where $p$ and $q$ represent the target and draft distributions, and the cost ratio $c = T_p/T_q$ between the two models. While neural drafts like EAGLE \citep{li2024eagle} or distillation methods \citep{zhou2024distillspecimprovingspeculativedecoding} improve $\alpha$ through alignment, they typically apply a uniform drafting policy regardless of the underlying text structure.

\paragraph{Context-Based and Adaptive Speculation.}
Beyond neural drafting, context-based methods like \textit{Prompt Lookup Decoding (PLD)} \citep{saxena2023prompt} use $n$-gram matching to propose candidates directly from the context. \textit{CopySpec} \citep{dumitru2025copyspec} introduced a heuristic switch between $n$-gram matching and neural drafting. However, these methods remain model-agnostic and fail at ``accidental repetitions.'' Recent work has explored statistical orchestration: \textit{BanditSpec} \citep{Hou2025BanditSpecAS} treats strategy selection as a multi-armed bandit problem, using UCB-based feedback on throughput to optimize drafting. Unlike these statistical ``black-box'' approaches that require noisy exploration phases, \methodname uses a ``white-box'' signal-probing internal states to identify copy-intent with high precision before speculation begins.

\paragraph{The Copy Mechanism in LLMs.}
We distinguish between memorization (training-data copying) and associative copying (contextual reproduction). The latter is driven by specific circuits known as \textit{induction heads} \citep{olsson2022context, elhage2021mathematical}. Conceptually, \methodname echoes the gating logic of \textit{Pointer-Generator Networks} \citep{See2017GetTT}, which used a learned scalar to switch between generation and pointing. However, we pivot this concept from generation-time vocabulary gating to inference-time speculation orchestration. Following evidence that task-relevant information is linearly encoded in $H^\ell$ \citep{park2024linear, hendel2023context, eisenstadt2025overclocking}, we hypothesize that copy-intent manifests as a linearly separable signal, allowing us to train lightweight probes as high-precision control signals for SD.

\vspace{-5pt}
\section{Method\label{sec:method}}
\vspace{-2pt}

\subsection{Motivation: The Efficiency Tax of Accidental Repetitions} 

We define \textit{accidental repetitions} as scenarios where the current context contains an $n$-gram overlap with the past, yet the model's target distribution $p$ does not align with the historical continuation. As illustrated in Figure~\ref{fig:Qualitative_accidental_vs_copy_intent}, these cases frequently arise from coincidental associative patterns (e.g., repeating the variable name ``$n$'' in a mathematical derivation) rather than structured reproduction. 

In such instances, indiscriminate copying incurs a significant \textit{speculation tax}: the system spends computational budget on draft proposals and parallel verification only to suffer a $0$-token acceptance rate. Our analysis shows that in reasoning-heavy tasks, these false-positives trigger over \textit{50\%} of all copy attempts. This motivates the need for a \textit{latent-aware trigger}-one that bypasses surface-level coincidences by directly querying the model's internal representations for genuine copy-intent.

\vspace{-3pt}
\subsection{Learning a Copy-Intent Probe\label{subsec:copy_probe}}
\vspace{-2pt}
Drawing on the insight that copying constitutes a distinct decoding regime, we develop a task-agnostic procedure to probe for copy-intent directly within the model's hidden states. Unlike heuristic triggers that monitor surface-level text, our probe identifies the underlying \textit{latent mode} of the transformer.

\vspace{-4pt}
\paragraph{Problem Formulation.}
\vspace{-3pt}
We formulate copy-intent detection as a binary classification task. Since surface-level repetitions are inherently noisy, we adopt a conservative proxy for our ground-truth training labels. We define a token position as a \textit{copy-mode} instance ($Y=1$) if it belongs to a verbatim sequence of at least $n_{\text{train}} = 5$ tokens that appears earlier in the context. This threshold ensures that our probe ignores incidental $n$-gram overlaps and focuses on sustained, deliberate copying behavior. All other positions are labeled as non-copy ($Y=0$). Our objective is not to claim that \textit{copy intent} is fully characterized by exact n-gram overlap, but rather to use longer verbatim spans as a practical indicator of an underlying latent \textit{copy mode}. Further discussion of this matter is in Appendix~\ref{sec:discussion}.


\vspace{-4pt}
\paragraph{Controlled Probing Dataset.}
\vspace{-3pt}

Because long repeated $n$-grams are sparse in standard pretraining corpora, we construct a \textit{calibrated stimulus set} to isolate the copy-circuit. We utilize Claude Sonnet \citep{anthropic_claude_sonnet} to generate 1,000 prompts that exhibit diverse repetition structures, ranging from structured code boilerplate to repetitive linguistic templates. By having the target model generate completions for these prompts, we extract a dataset $\mathcal{D}_\ell$ of hidden representations $H_j^{(\ell)}$ paired with their corresponding copy-labels $Y_j$:
\begin{equation}
    \mathcal{D}_\ell := \{ (H_{i,j}^{(\ell)}, Y_{i,j}) \mid i \in [1000], j \in \{M_i, \dots, N_i\} \} \,,
\end{equation}
where $M_i$ denotes the start of the decoded response. This procedure provides hundreds of thousands of labeled token-level instances, ensuring a high signal-to-noise ratio for probe training. Appendix~\ref{app:CopyDataset} provides representative prompts from the dataset, illustrating their varying degrees of copy-inducing behavior.

\paragraph{Probe Architecture and Training.}
Consistent with the hypothesis that task-relevant features are linearly encoded in the residual stream \citep{park2024linear}, we implement the probe as a \textit{strictly linear projection}. For each layer $\ell$, we train an independent probe $\mathcal{PR}_\ell$ that operates directly on the hidden representation $H_j^{(\ell)} \in \mathbb{R}^d$ without a bias term:
\begin{equation}
    p_\ell(j) = \sigma(w_\ell^\top H_j^{(\ell)}) \,,
\end{equation}
where $p_\ell(j)$ represents the probability that the model is in copy-mode. Probes are trained using binary cross-entropy loss with $\ell_2$ regularization. This minimal architecture ensures that probe predictions are purely a function of the vector direction in the latent space, adding negligible computational overhead. Empirically, we find that probes trained on representations extracted after the \textit{attention sublayer} consistently outperform those from the MLP sublayers, reinforcing the connection to attention-driven induction circuits.

\vspace{-3pt}
\paragraph{Layer and Threshold Selection.}
After training probes across all $L$ layers, we select a single optimal layer using a greedy F1-score maximization on a held-out validation set. This identifies the specific depth at which copy-intent is most linearly separable, typically in the intermediate layers (e.g., layer 14 for Llama-3.1-8B). We then fix a decision threshold $\tau$ by maximizing the F1-score on the same validation split. At inference time, \methodname enables copy-based drafting only when the probe score $\mathcal{PR}(H_j^{(\ell)}) \geq \tau$, providing a high-precision gate that filters out accidental repetitions.

\vspace{-7pt}
\subsection{\methodname: Copy-Intent-Aware SD\label{subsec:SwitchSD}}
\vspace{-3pt}

Building upon the selected probe $\mathcal{PR}$, \methodname employs a dynamic speculative framework that switches between neural drafting and context-copying strategies. At each decoding step, the system extracts the hidden representation $H_i^{(\ell)}$ and queries the probe to determine the optimal drafting regime (see Figure~\ref{fig:specSwtich}).

\paragraph{Copy-Intent Decision Rule.}
Let $i$ denote the current token position. Copy-based speculation is enabled if and only if the latent probe score exceeds the threshold $\tau$, \textit{and} the current context provides structural evidence for a repetition:
\begin{equation}\label{eq:decision_rule}
    \mathcal{PR}(H_i^{(\ell)}) \ge \tau \quad \text{where} \quad \mathcal{PR}(H_i^{(\ell)}) = \sigma(w^\top H_i^{(\ell)}) \,.
\end{equation}
To suppress accidental short-prefix matches, we further require that the $\kappa$-gram ending at the current token $(x_{i-\kappa+1}, \dots, x_i)$ matches at least one prior occurrence in the context (we set $\kappa=5$). If Eq.~\ref{eq:decision_rule} holds but no $\kappa$-gram match is found, \methodname falls back to neural drafting, ensuring that speculation is only triggered when both internal intent and external structure align.

\vspace{-3pt}

\paragraph{Copy-Based Speculative Generation.}
When the copy-regime is triggered, \methodname retrieves up to $W$ candidate continuations, each of length $D$, from the context. These candidates are ranked by recency, prioritizing the most recent linguistic patterns. This yields a candidate set of plausible verbatim continuations without requiring any additional draft model parameters.

\paragraph{Parallel Verification with Block-Triangular Attention.}
To verify these $W$ candidates efficiently, we adopt a \textit{block-triangular attention} pattern, enabling parallel validation in a single target-model forward pass. We concatenate the $W$ candidates sequentially after the anchor position. The attention mask is structured such that tokens within a specific candidate $w \in [W]$ can attend to: (i) the full original prefix, and (ii) preceding tokens within candidate $w$ itself, while remaining isolated from other candidates. This allows the target model to verify all hypotheses simultaneously. \methodname then selects the longest validated continuation for the next decoding step.

\paragraph{Non-Copy Speculative Tokens.}
In generative regimes where copy-intent is not detected, \methodname delegates drafting to traditional SD mechanisms. To demonstrate the framework's flexibility, we evaluate \methodname using both vanilla \textit{Speculative Sampling (SPS)} and the state-of-the-art \textit{EAGLE3} \citep{li2024eagle}. This "orchestrator-first" design proves that signal-aware switching provides benefits that are orthogonal to the specific choice of a neural draft model.

\paragraph{KV Cache Synchronization.}
Orchestrating between copy-based and neural-based paths requires careful coordination of Key-Value (KV) caches. When a copy-based sequence is accepted, the draft model's KV cache may lag behind. \methodname resolves this via \textit{lazy propagation}: accepted tokens are passed through the draft model during the subsequent forward pass. By merging this synchronization with the next drafting step, the synchronization cost is effectively \textit{amortized}, ensuring that both the target and draft caches remain coherent with minimal latency overhead.

\paragraph{Overall Framework.}
\methodname acts as an orchestration layer over SD: at each step, it dynamically selects between copy-based and draft-based speculation, while delegating final verification to the target model. This design preserves correctness guarantees and is fully compatible with both classical speculative sampling and state-of-the-art methods such as \textit{EAGLE3}.

\vspace{-9pt}
\section{Experiments}
\vspace{-5pt}
This section presents a systematic evaluation of \methodname. We begin by reporting main results across multiple benchmarks, comparing \methodname against heuristic, statistical, and neural baselines (\S\ref{subsec:mainRes}). We then provide a detailed model analysis to characterize how latent signals improve acceptance lengths (\S\ref{subsec:modelAnlyze}). Finally, we justify our design choices through dedicated ablation studies 
(\S\ref{subsec:ablation}). Additional experiments analyzing the impact of temperature sampling are provided in App.~\ref{app:impact_of_temp}, showing that \methodname maintains SoTA performance even at high temperatures, and additional comparisons against context-exploitation baselines are presented in App.~\ref{app:additional_baselines}.

\begin{table*}[t]
\centering
\scriptsize
\setlength{\tabcolsep}{3pt}
\caption{Throughput \& speedup (relative to vanilla) across datasets and models. Each model block reports Tok/s, Speedup, and the copy/EAGLE3 iteration breakdown as \% iterations and mean acceptance length.}
\renewcommand{\arraystretch}{1.15}
\resizebox{\textwidth}{!}{%
\begin{tabular}{@{}l@{~~}l@{~}
c@{~}c@{~}c@{~}c@{~}c@{~}c@{~}
c@{~}c@{~}c@{~}c@{~}c@{~}c@{~}
c@{~}c@{~}c@{~}c@{~}c@{~}c@{}}
\toprule
\multirow{3}{*}{Dataset} & \multirow{3}{*}{Method}
& \multicolumn{6}{c}{\textbf{LLaMA-3.1-8B-Instruct}}
& \multicolumn{6}{c}{\textbf{LLaMA-3.3-70B-Instruct}}
& \multicolumn{6}{c}{\textbf{Qwen3-8B}} \\
\cmidrule(lr){3-8}\cmidrule(lr){9-14}\cmidrule(lr){15-20}
& & \multirow{2}{*}{Tok/s} & \multirow{2}{*}{Speedup}
  & \multicolumn{2}{c}{Copy} & \multicolumn{2}{c}{EAGLE3}
  & \multirow{2}{*}{Tok/s} & \multirow{2}{*}{Speedup}
  & \multicolumn{2}{c}{Copy} & \multicolumn{2}{c}{EAGLE3}
  & \multirow{2}{*}{Tok/s} & \multirow{2}{*}{Speedup}
  & \multicolumn{2}{c}{Copy} & \multicolumn{2}{c}{EAGLE3} \\
\cmidrule(lr){5-6}\cmidrule(lr){7-8}\cmidrule(lr){11-12}\cmidrule(lr){13-14}\cmidrule(lr){17-18}\cmidrule(lr){19-20}
& & & & \% & mean & \% & mean & & & \% & mean & \% & mean & & & \% & mean & \% & mean \\
\midrule

\multirow{6}{*}{\textbf{CNN/DM}}
& Vanilla & 38.24 & 1.00$\times$ & -- & -- & -- & -- & 11.46 & 1.00$\times$ & -- & -- & -- & -- & 25.67 & 1.00$\times$ & -- & -- & -- & -- \\
& PLD & 50.70 & 1.33$\times$ & 100\% & 1.51 & -- & -- & -- & -- & -- & -- & -- & -- & -- & -- & -- & -- & -- & -- \\
& BanditSpec & 63.60 & 1.66$\times$ & 21.0\% & 0.44 & 79.0\% & 2.42 & 11.60 & 1.01$\times$ & 96.0\% & 0.34 & 4.0\% & 0.89 & 49.80 & 1.93$\times$ & 2.0\% & 0.05 & 98.0\% & 2.15 \\
& CopySpec & 66.39 & 1.74$\times$ & 14.2\% & 1.50 & 85.8\% & 2.32 & 15.74 & 1.37$\times$ & 10.0\% & 1.01 & 90.0\% & 0.98 & 50.04 & 1.95$\times$ & 5.7\% & 0.84 & 94.3\% & 2.19 \\
& EAGLE3 & 70.84 & 1.85$\times$ & -- & -- & 100\% & 2.60 & 15.71 & 1.37$\times$ & -- & -- & 100\% & 1.03 & 51.44 & 2.00$\times$ & -- & -- & 100\% & 2.20 \\
& \textbf{SwitchSD} & \textbf{77.00} & \textbf{2.01$\times$} & 10.6\% & 2.84 & 89.4\% & 2.46 & \textbf{16.22} & \textbf{1.42$\times$} & 5.4\% & 1.96 & 94.6\% & 1.04 & \textbf{53.23} & \textbf{2.07$\times$} & 2.1\% & 1.97 & 97.9\% & 2.20 \\
\midrule

\multirow{6}{*}{\textbf{Math500}}
& Vanilla & 38.75 & 1.00$\times$ & -- & -- & -- & -- & 11.80 & 1.00$\times$ & -- & -- & -- & -- & 27.11 & 1.00$\times$ & -- & -- & -- & -- \\
& PLD & 53.10 & 1.37$\times$ & 100\% & 1.54 & -- & -- & -- & -- & -- & -- & -- & -- & -- & -- & -- & -- & -- & -- \\
& BanditSpec & 72.50 & 1.87$\times$ & 5.0\% & 0.30 & 95.0\% & 2.58 & 11.70 & 0.99$\times$ & 91.0\% & 0.28 & 9.0\% & 0.66 & 59.02 & 2.18$\times$ & 1.0\% & 0.15 & 99.0\% & 2.58 \\
& CopySpec & 74.00 & 1.91$\times$ & 30.3\% & 2.09 & 69.7\% & 2.34 & 15.82 & 1.34$\times$ & 19.0\% & 1.50 & 81.0\% & 0.85 & 55.28 & 2.04$\times$ & 17.4\% & 1.00 & 82.6\% & 2.58 \\
& EAGLE3 & 79.38 & 2.05$\times$ & -- & -- & 100\% & 2.87 & 15.87 & 1.35$\times$ & -- & -- & 100\% & 0.96 & 59.77 & 2.20$\times$ & -- & -- & 100\% & 2.65 \\
& \textbf{SwitchSD} & \textbf{88.74} & \textbf{2.29$\times$} & 17.9\% & 5.39 & 82.1\% & 2.41 & \textbf{17.41} & \textbf{1.48$\times$} & 8.9\% & 4.45 & 91.1\% & 0.82 & \textbf{62.67} & \textbf{2.31$\times$} & 9.3\% & 2.68 & 90.7\% & 2.57 \\
\midrule

\multirow{6}{*}{\textbf{HumanEval}}
& Vanilla & 38.30 & 1.00$\times$ & -- & -- & -- & -- & 11.73 & 1.00$\times$ & -- & -- & -- & -- & 26.02 & 1.00$\times$ & -- & -- & -- & -- \\
& PLD & 53.00 & 1.38$\times$ & 100\% & 1.59 & -- & -- & -- & -- & -- & -- & -- & -- & -- & -- & -- & -- & -- & -- \\
& BanditSpec & 75.00 & 1.96$\times$ & 28.0\% & 0.45 & 72.0\% & 3.25 & 12.70 & 1.08$\times$ & 75.0\% & 0.30 & 25.0\% & 1.11 & 64.63 & 2.49$\times$ & 1.0\% & 0.13 & 99.0\% & 2.88 \\
& CopySpec & 77.72 & 2.03$\times$ & 27.3\% & 1.83 & 72.7\% & 2.82 & 19.17 & 1.63$\times$ & 14.4\% & 1.72 & 85.6\% & 1.32 & 61.92 & 2.38$\times$ & 15.8\% & 1.01 & 84.2\% & 2.92 \\
& EAGLE3 & 87.50 & 2.28$\times$ & -- & -- & 100\% & 3.44 & 19.52 & 1.66$\times$ & -- & -- & 100\% & 1.44 & 66.52 & 2.56$\times$ & -- & -- & 100\% & 2.97 \\
& \textbf{SwitchSD} & \textbf{98.63} & \textbf{2.58$\times$} & 14.2\% & 5.88 & 85.8\% & 2.93 & \textbf{20.46} & \textbf{1.74$\times$} & 7.0\% & 4.65 & 93.0\% & 1.27 & \textbf{67.98} & \textbf{2.61$\times$} & 7.9\% & 3.16 & 92.1\% & 2.91 \\
\bottomrule
\end{tabular}%
}
\label{tab:all-models-pivoted}
\vspace{-12pt}
\end{table*}

\vspace{-6pt}
\subsection{Main Result\label{subsec:mainRes}}
\vspace{-4pt}

We evaluate \methodname across three primary domains: coding (HumanEval~\citep{chen2021codex}), mathematical reasoning (Math500~\citep{hendrycks2021measuring}), and summarization (CNN/DailyMail~\citep{Hermann2015TeachingMT}). Performance is measured by decoding throughput (tokens per second) and the resulting speedup relative to vanilla autoregressive decoding. 

\vspace{-5pt}
\paragraph{Baselines \& Models.}
We compare \methodname against three classes of speculative decoding:
(i) \textit{Heuristic-based:} \textit{PLD} \citep{saxena2023prompt} (naive $n$-gram search) and \textit{CopySpec} \citep{dumitru2025copyspec} (prefix-based switching). 
(ii) \textit{Neural-only:} \textit{EAGLE3} \citep{li2025eagle}, the current state-of-the-art in neural speculative drafting.
(iii) \textit{Statistical-Adaptive:} \textit{BanditSpec} \citep{Hou2025BanditSpecAS}, which uses a UCB-based bandit to switch between strategies based on throughput rewards. For model selection, we conduct experiments using models that are already supported by EAGLE3. In particular, we evaluate models from the LLaMA-3~\citep{grattafiori2024llama} and Qwen-3~\citep{yang2025qwen3} families.


\begin{wraptable}{r}{0.46\linewidth}
\vspace{-12pt}
\centering
\scriptsize
\setlength{\tabcolsep}{3pt}
\caption{Decoding throughput and speedup for Qwen3-8B (8-bit) using SPS with a Qwen3-0.6B draft model. Copy and SPS statistics report \% iterations and mean acceptance length.}
\vspace{-2pt}
\renewcommand{\arraystretch}{1.1}
\begin{tabular}{@{}l@{~}l@{~}c@{~}c@{~}c@{~}c@{~}c@{~}c@{}}
\toprule
Dataset & Method & Tok/s & Speedup
& \multicolumn{2}{c}{Copy}
& \multicolumn{2}{c}{SPS} \\
\cmidrule(lr){5-6}\cmidrule(lr){7-8}
& & & & \% & mean & \% & mean \\
\midrule

\multirow{4}{*}{\textbf{CNN/DM}}
& Vanilla    & 7.53  & 1.00$\times$ & -- & -- & -- & -- \\
& CopySpec   & 10.30 & 1.37$\times$ & 5.8 & 1.77 & 94.2 & 2.46 \\
& SPS        & 9.74  & 1.29$\times$ & -- & -- & 100 & 2.72 \\
& \textbf{SwitchSD}
              & \textbf{10.43} & \textbf{1.38$\times$}
              & 3.2 & 2.79 & 96.8 & 2.47 \\
\midrule

\multirow{4}{*}{\textbf{Math500}}
& Vanilla    & 7.84  & 1.00$\times$ & -- & -- & -- & -- \\
& CopySpec   & 13.18 & 1.68$\times$ & 17.7 & 2.05 & 82.3 & 3.09 \\
& SPS        & 13.47 & 1.72$\times$ & -- & -- & 100 & 3.77 \\
& \textbf{SwitchSD}
              & \textbf{13.85} & \textbf{1.77$\times$}
              & 8.6 & 4.08 & 91.4 & 3.10 \\
\midrule

\multirow{4}{*}{\textbf{HumanEval}}
& Vanilla    & 7.87  & 1.00$\times$ & -- & -- & -- & -- \\
& CopySpec   & 12.31 & 1.56$\times$ & 14.3 & 2.08 & 85.7 & 2.88 \\
& SPS        & 12.35 & 1.57$\times$ & -- & -- & 100 & 3.39 \\
& \textbf{SwitchSD}
              & \textbf{13.09} & \textbf{1.66$\times$}
              & 7.2 & 4.37 & 92.8 & 2.89 \\
\bottomrule
\end{tabular}
\label{tab:qwen8b-sps-onecol}
\vspace{-16pt}
\end{wraptable}

\paragraph{Throughput and Speedup.}
Across all 9 configurations in Table~\ref{tab:all-models-pivoted}, \methodname consistently achieves the highest throughput, outperforming all baselines. On average, we achieve speedups of \textit{2.31$\times$} on HumanEval, \textit{2.03$\times$} on Math500, and \textit{1.83$\times$} on CNN/DM. Notably, \methodname provides a consistent performance gain over EAGLE3 (up to 15\%), demonstrating that signal-aware context-copying is orthogonal and complementary to the strongest neural speculators. These trends hold even when using a weaker draft model (SPS) on Qwen3-8B (Table~\ref{tab:qwen8b-sps-onecol}), where \methodname continues to dominate.
\vspace{-4pt}
\paragraph{Internal Signals vs. Statistical Exploration.}
A key finding is our performance relative to \textit{BanditSpec}. While BanditSpec attempts to adaptively switch strategies, it suffers from a significant \textit{exploration penalty}. As a black-box optimizer, it must "explore" the copy strategy in generative regions to collect throughput statistics. On Llama-3.3-70B, this leads to a near-total collapse in efficiency (e.g., 0.99$\times$ on Math500), where it over-selects copying (91\% of iterations) despite a catastrophic mean acceptance length of only 0.28. 

In contrast, \methodname achieves a \textit{1.48$\times$} speedup on the same task. By leveraging internal hidden-state signals, \methodname identifies copy-intent \textit{before} drafting, allowing it to be more selective (triggering only 8.9\% of the time) but achieving a much higher yield (4.45 mean acceptance). This isolates the core benefit of our approach: internal representation probing provides a deterministic control signal that far exceeds the efficiency of reward-based statistical feedback.

\paragraph{High-Yield Speculation.}
The breakdown of acceptance lengths in Table~\ref{tab:all-models-pivoted} reveals that \methodname achieves its gains not through frequency, but through \textit{precision}. Compared to CopySpec, \methodname triggers copy-based drafting less often but attains significantly longer accepted runs, for instance, 5.88 vs. 1.83 on Llama-3.1-8B HumanEval. This confirms that our probe successfully identifies the high-yield ``copy-regimes'' of the model while suppressing the noisy ``accidental repetitions'' that plague heuristic and statistical methods.

\paragraph{Robustness Across SD Architectures.}
To evaluate whether the benefits of \methodname are tied to specific neural SD architectures, we report results for Qwen3-8B using standard \textit{Speculative Sampling (SPS)} with a 0.6B draft model in Table~\ref{tab:qwen8b-sps-onecol}. Unlike EAGLE3, which predicts future feature states, SPS relies on token-level sampling from a smaller model. 

Despite this shift in the underlying neural speculator, \methodname consistently achieves the highest throughput across all benchmarks. For example, on HumanEval, it increases throughput to 13.09 Tok/s, outperforming both the base SPS (12.35) and the heuristic-based CopySpec (12.31). These results confirm that signal-aware orchestration is \textit{draft-agnostic}; the ability to identify a ``copy-regime'' via internal representations provides a significant efficiency layer that is orthogonal to the strength or type of the neural draft model.

\paragraph{Precision Advantage Carries Across SD Architectures.}
A comparative analysis of the iteration statistics in Table~\ref{tab:qwen8b-sps-onecol} reveals the fundamental mechanism behind our gains. Across all datasets, \methodname invokes copying significantly more \textit{selectively} than the heuristic CopySpec, yet achieves \textit{longer accepted spans}. This ``less is more'' pattern is the mathematical signature of our latent copy-intent probe: (i)\textit{HumanEval:} \methodname triggers context-copying in only 7.2\% of steps (vs. 14.3\% for CopySpec), yet yields a mean acceptance length of 4.37 (vs. 2.08). (ii) \textit{Math500:} We trigger copying in only 8.6\% of steps (vs. 17.7\%), but accept 4.08 tokens on average (vs. 2.05).

By filtering out the ``accidental repetitions'' where $n$-gram overlaps exist but the model's actual intent remains generative, \methodname ensures that context-copying is only utilized when it can provide high-yield leaps. This precision allows the neural draft model to focus on the generative regions where it excels, leading to a more efficient division of labor in the decoding pipeline.

\paragraph{Task-Specific Performance.}
The performance gains are most pronounced in structured domains like coding and mathematics. In these tasks, the model transitions between rigid, repetitive logic (copy-heavy) and novel reasoning (generative). \methodname successfully navigates these transitions by ``listening'' to the model's internal state. In contrast, on CNN/DailyMail, where the repetition is naturally weaker, the gains are more modest but still favor the signal-aware approach. Together, these results demonstrate that \methodname acts as a robust, universal orchestrator that optimizes the interplay between contextual and neural speculation.

\vspace{-3pt}
\subsection{Model Analysis\label{subsec:modelAnlyze}}

\subsubsection{Regime Separation and Acceptance Yield}
\begin{wrapfigure}{r}{0.6\linewidth}
    \vspace{-15pt}
    \centering
\includegraphics[width=1.0\linewidth]
    {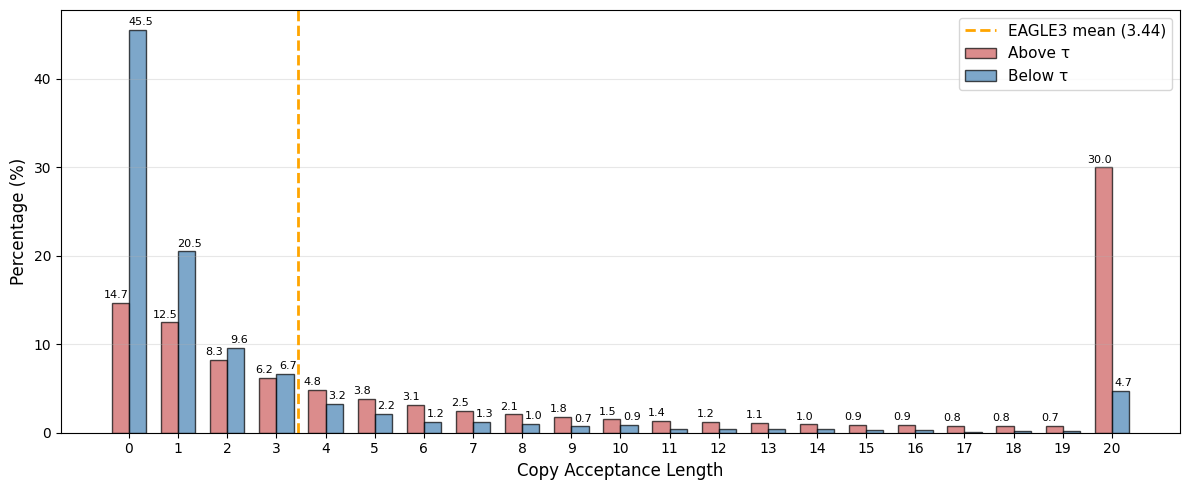}
    \vspace{-7pt}
    \caption{\small \textbf{Copy acceptance length conditioned on probe prediction.} Distribution of the number of speculative tokens accepted from the copy-based draft partitioned by the probe’s prediction of copy intent. Red bars correspond to steps where copy intent is identified (probe active), while blue bars correspond to steps where copy intent is not predicted (probe inactive).}
    \label{fig:copy_acc_distribution}
    \vspace{-13pt}
\end{wrapfigure}

In this section, we provide a fine-grained analysis of \methodname's internal mechanics, demonstrating how latent-signal probing successfully partitions decoding into two distinct mathematical regimes.

To quantify the probe's ability to identify high-yield sequences, we perform a post-hoc analysis on \textsc{HumanEval} using \texttt{Llama-3.1-8B-Instruct}. We execute both the context-copying and EAGLE3 paths at every step, partitioning the results based on the probe's prediction. Figure~\ref{fig:copy_acc_distribution} illustrates the resulting probability mass functions for speculative acceptance lengths. The distributions reveal a near-total separation of decoding modes. When the probe is \textit{inactive} (blue), the copy-based draft suffers from a heavy concentration of short acceptances ($0$--$2$ tokens), representing the ``accidental repetition'' regime where copying is inefficient. Conversely, when the probe is \textit{active} (red), the distribution shifts dramatically toward high-yield runs. Most notably, \textit{30.0\%} of trajectories reach the maximum speculative length of 20 tokens when the probe is active, compared to a mere \textit{4.7\%} when it is inactive. This \textit{6.3$\times$} increase in maximum-length yield confirms that our probe identifies a latent state of high predictability that surface-level heuristics cannot detect. These results suggest that the probe captures an internal copy-intent signal rather than simply detecting surface-level repetition patterns. We provide a broader discussion of this interpretation and its limitations in Appendix~\ref{sec:discussion}.

\subsubsection{Impact on Optimal Lookahead ($\gamma$)}

\begin{wraptable}{r}{0.44\linewidth}
\vspace{-10pt}
\centering
\scriptsize
\caption{\small Best draft lookahead $\gamma$ selected on the evaluation set for EAGLE3 and \methodname.}
\vspace{-4pt}
\setlength{\tabcolsep}{4pt}
\begin{tabular}{lccc}
\toprule
Approach & LLaMA-3.1-8B & Qwen3-8B & LLaMA-3.3-70B \\
\midrule
Vanilla & 7 & 5 & 7 \\
SwitchSD & 6 & 5 & 5 \\
\bottomrule
\end{tabular}
\vspace{0.3em}
\label{tab:gamma_results}
\vspace{-9pt}
\end{wraptable}

The partitioning of decoding into ``copy'' and ``generative'' regimes has significant implications for the neural draft's optimal lookahead, $\gamma$. Typically, $\gamma$ is tuned based on the average acceptance rate across all tokens. However, context-copying tokens are inherently easier to speculate, which artificially inflates the global acceptance rate. 

By routing these high-probability tokens through the copy path, \methodname removes the ``easy'' examples from the neural draft's workload. As a result, the remaining tokens assigned to the neural speculator are harder on average, leading to the slightly lower optimal $\gamma$ values observed in Table~\ref{tab:gamma_results} (e.g., a shift from $7$ to $5$ for Llama-3.3-70B). 

This separation enables a critical \textit{system-level optimization}: instead of over-speculating on generative tokens with a high global $\gamma$, \methodname allows for targeted tuning. By using a shorter, more appropriate lookahead for the neural draft and a high-yield span for the copy draft, we eliminate computational waste and maximize the throughput of both paths. This refined resource allocation is a primary driver of the speedups reported in \S\ref{subsec:mainRes}.

\vspace{-5pt}
\subsection{Ablation Studies\label{subsec:ablation}}
We conduct a series of ablations to isolate the impact of our training data, representational placement, and decision thresholds on \methodname's performance.
%

\begin{wrapfigure}{r}{0.6\linewidth}
    \vspace{-10pt}
  \centering
  \includegraphics[width=1.0\linewidth]{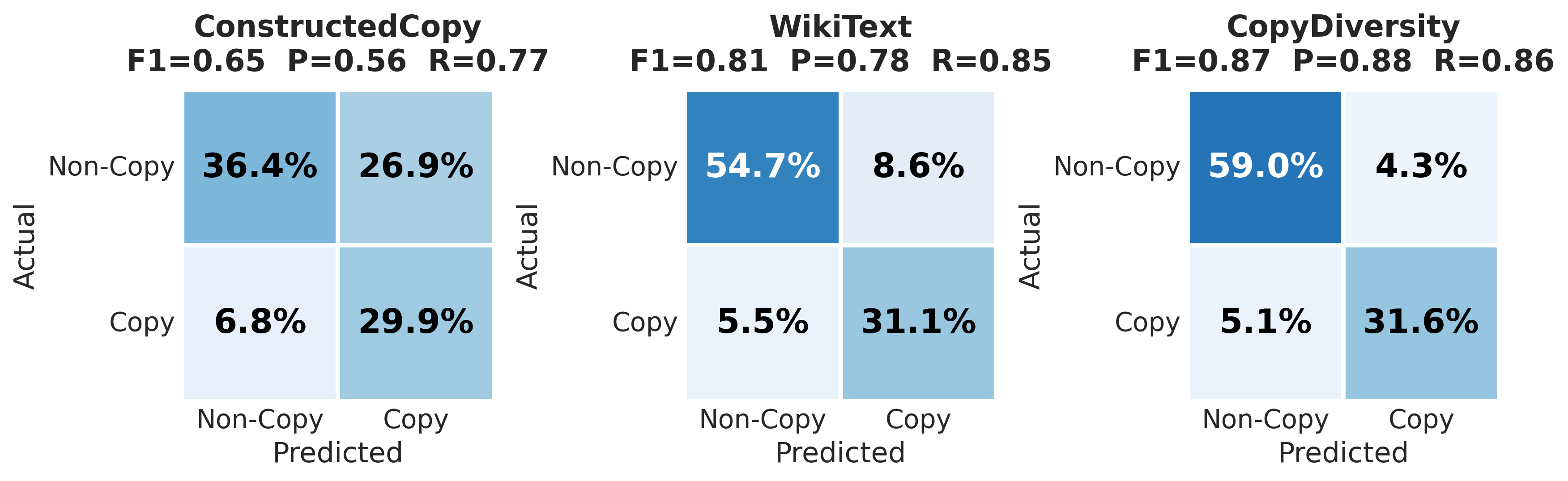}
  \caption{\small Confusion matrices for copy-intent probes evaluated on MT-Bench at their selected decision thresholds.
  Each cell reports the percentage of tokens from the total evaluation set.}
  \label{fig:mtbench_confusion}
\end{wrapfigure}

\subsubsection{Importance of Intent-Driven Training Data}
To evaluate our synthetic data construction, we compare the efficacy of probes trained on three distinct data distributions: (i) \textit{ConstructedCopy}, a non-semantic dataset of repeated random token spans; (ii) \textit{WikiText-103}, representing natural linguistic repetition without explicit prompt-completion structure; and (iii) \textit{CopyDiversity} (our method), which utilizes model-generated completions across diverse repetition templates.

As shown in Figure~\ref{fig:mtbench_confusion}, the non-semantic \textit{ConstructedCopy} yields the weakest performance ($F1=0.65$), confirming that copy-intent is not merely a structural phenomenon but is tied to the model's semantic processing. While \textit{WikiText-103} improves performance to $0.81$, it remains significantly below our full method ($0.87$). The confusion matrices illustrate this gap: probes trained on natural text or random spans suffer from higher false-positive rates on \textsc{MT-Bench}, likely failing to distinguish between coincidental reuse and intentional reproduction. These results underscore that high-precision probing requires \textit{intent-driven} supervision that aligns with the target model's actual 
behavior.

\begin{wrapfigure}{r}{0.5\linewidth}
    \centering
    \includegraphics[width=1.0\linewidth]{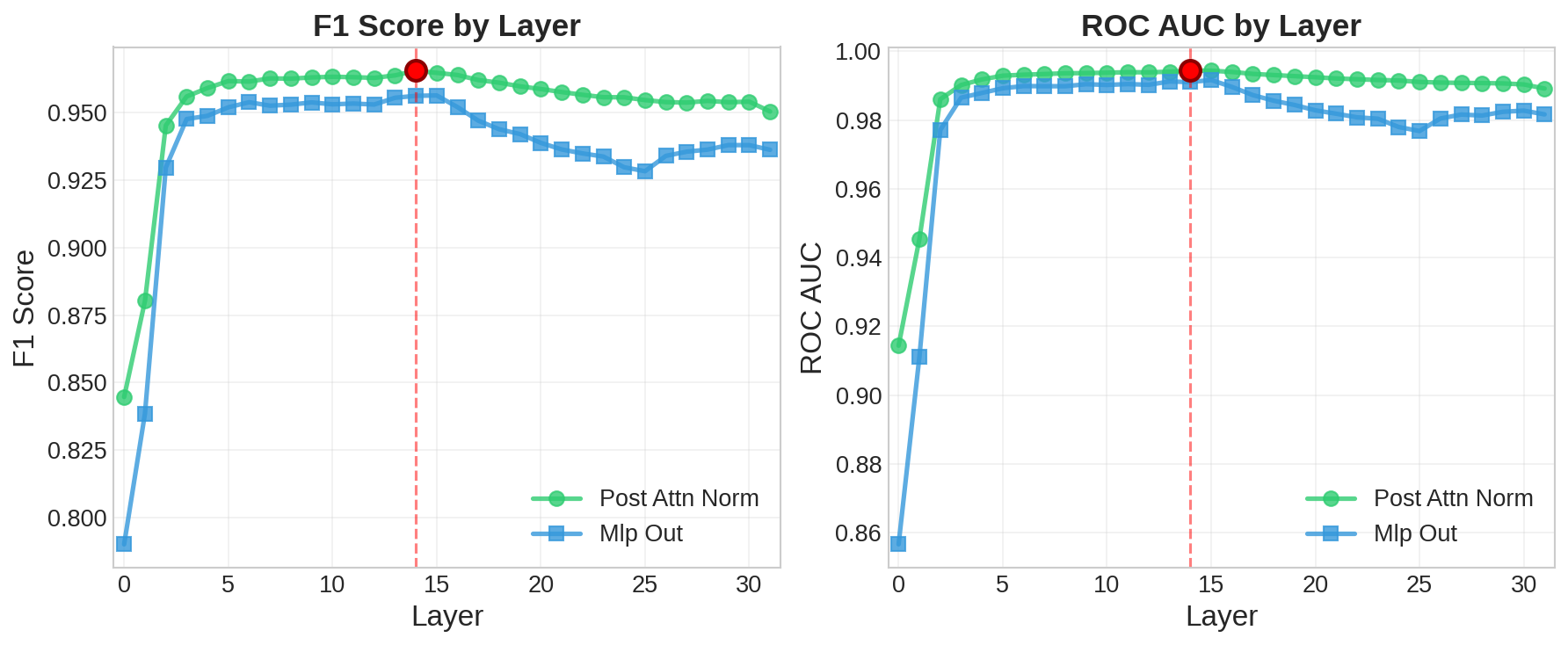}
    \caption{\small \textbf{Position and layer selection for probing.} F1 (left) and ROC–AUC (right) of probes trained on representations extracted after the attention sublayer (green) or after the MLP sublayer (blue) across transformer layers.}
    \label{fig:layer_and_placment_select}
\end{wrapfigure}

\vspace{15pt}

\subsubsection{Architectural Placement: Layer and Sublayer Selection}

A critical design choice is the selection of internal representations. We evaluate probes trained on representations extracted from every layer at two positions: (i) after the self-attention sublayer ($X'$) and (ii) after the MLP sublayer ($Y$).

Figure~\ref{fig:layer_and_placment_select} reveals two key insights. First, probes trained on \textit{attention representations} consistently outperform those from MLP sublayers. This result aligns with our hypothesis in \S2, suggesting that copy-intent is primarily carried by attention-driven induction circuits rather than feed-forward transformations. Second, the signal strength peaks in the \textit{intermediate layers} (peaking at layer 14 for Llama-3.1-8B), while declining in the final layers as representations transform into vocabulary logits. This validates our choice of intermediate attention-based hidden states as the optimal control signal.

\begin{wrapfigure}{r}{0.5\linewidth}
    \vspace{-10pt}
    \centering
    \includegraphics[width=1.0\linewidth]{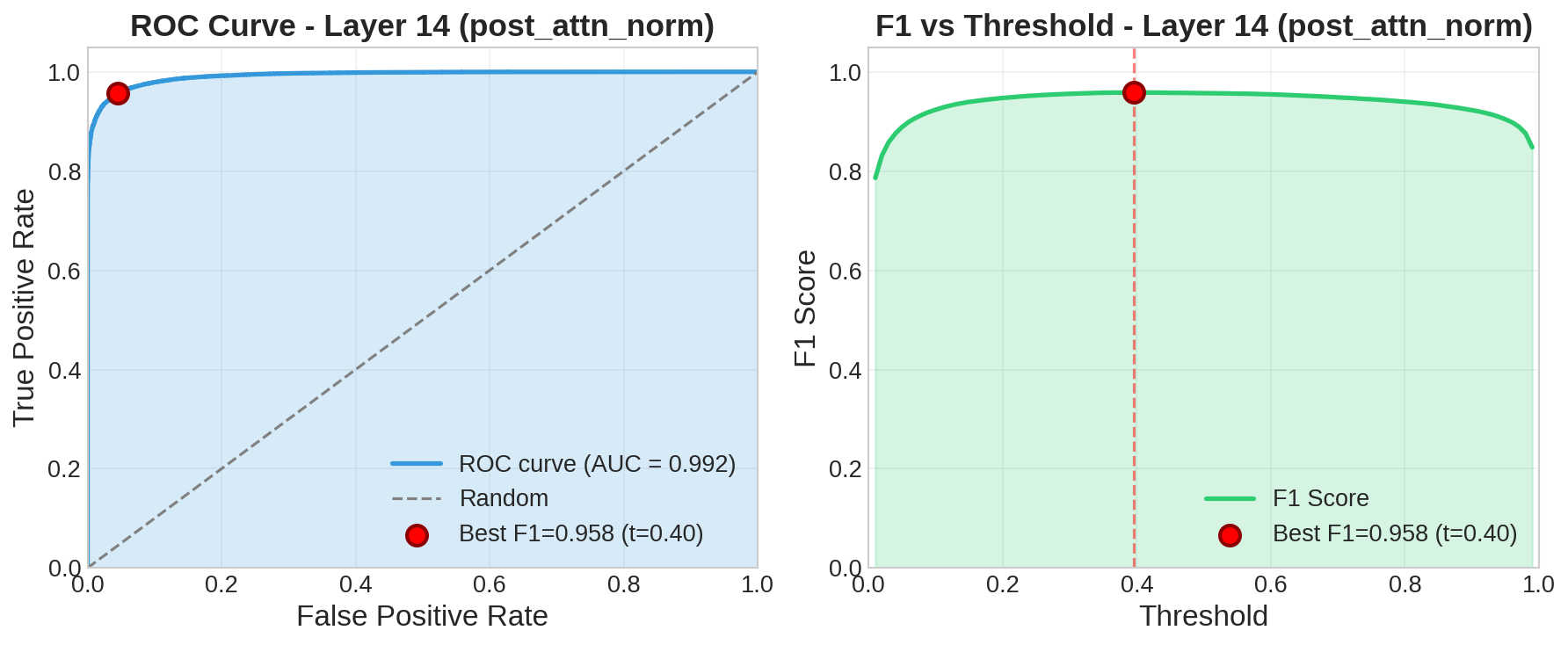}
    \caption{\small \textbf{Threshold selection for probing.} F1 score as a function of the decision threshold (right) and the corresponding ROC curve (left) for probes trained on attention representations at layer 14.}
    \label{fig:threshold_selection_fig}
    \vspace{-1pt}
\end{wrapfigure}

\subsubsection{Threshold Selection and Linear Separability}
Finally, we analyze the impact of the decision threshold $\tau$. Figure~\ref{fig:threshold_selection_fig} (left) demonstrates that an F1-optimal threshold lies near $0.4$ for the Llama-3.1-8B model. 

The corresponding ROC curve (Figure~\ref{fig:threshold_selection_fig}, right) exhibits an \textit{AUC $> 0.99$}, indicating near-perfect linear separability of the copy-mode. This exceptional clarity suggests that copy-intent is a \textit{first-order feature} in the model’s residual stream. Such high separability allows \methodname to operate as a high-precision gate, achieving the significant speedups reported in \S\ref{subsec:mainRes} with negligible risk of false-positive copy triggers.

\section{Conclusions}
We introduced \methodname, a framework that establishes a new direction for \textit{latent-aware orchestration} in speculative decoding. By moving beyond surface-level $n$-gram heuristics and statistical bandits, we demonstrate that a model's internal representations provide a high-precision, ``white-box'' control signal for switching between decoding regimes. Across diverse model families (Llama-3, Qwen) and benchmarks (Math500, HumanEval, CNN/Dailymail), \methodname consistently delivers significant throughput gains, outperforming state-of-the-art neural speculators like EAGLE3 by up to 15\% while preserving exact generation quality. Our findings suggest that the future of efficient LLM inference lies in \textit{model-aware systems} that exploit the intrinsic representational states of the transformer to optimize the division of labor between neural and contextual speculation.

Furthermore, our binary copy-intent probe could be extended into a \textit{multi-modal orchestrator} capable of predicting optimal speculation lengths ($\gamma$) or selecting between multiple specialized draft models based on latent task-type signals. Finally, we believe that deeper mechanistic interpretability of the ``copy-circuits'' identified in this work could lead to even more refined probes, further closing the gap between the theoretical understanding of transformer internals and the practical requirements of low-latency inference.

\newpage

\bibliographystyle{plainnat}
\bibliography{example_paper}

@misc{saxena2023prompt,
    title = {Prompt Lookup Decoding},
    author = {Apoorv Saxena},
    year = {2023},
    month = {November},
    url = {https://github.com/apoorvumang/prompt-lookup-decoding/}
}

@article{dumitru2025copyspec,
  title={CopySpec: Accelerating LLMs with Speculative Copy-and-Paste Without Compromising Quality},
  author={Dumitru, Razvan-Gabriel and Yang, Minglai and Yadav, Vikas and Surdeanu, Mihai},
  journal={arXiv preprint arXiv:2502.08923},
  year={2025}
}

@inproceedings{leviathan2023fast,
  title={Fast inference from transformers via speculative decoding},
  author={Leviathan, Yaniv and Kalman, Matan and Matias, Yossi},
  booktitle={International Conference on Machine Learning},
  pages={19274--19286},
  year={2023},
  organization={PMLR}
}

@article{li2024eagle,
  title={Eagle: Speculative sampling requires rethinking feature uncertainty},
  author={Li, Yuhui and Wei, Fangyun and Zhang, Chao and Zhang, Hongyang},
  journal={arXiv preprint arXiv:2401.15077},
  year={2024}
}

@article{chen2021codex,
  title={Evaluating Large Language Models Trained on Code},
  author={Mark Chen and Jerry Tworek and Heewoo Jun and Qiming Yuan and Henrique Ponde de Oliveira Pinto and Jared Kaplan and Harri Edwards and Yuri Burda and Nicholas Joseph and Greg Brockman and Alex Ray and Raul Puri and Gretchen Krueger and Michael Petrov and Heidy Khlaaf and Girish Sastry and Pamela Mishkin and Brooke Chan and Scott Gray and Nick Ryder and Mikhail Pavlov and Alethea Power and Lukasz Kaiser and Mohammad Bavarian and Clemens Winter and Philippe Tillet and Felipe Petroski Such and Dave Cummings and Matthias Plappert and Fotios Chantzis and Elizabeth Barnes and Ariel Herbert-Voss and William Hebgen Guss and Alex Nichol and Alex Paino and Nikolas Tezak and Jie Tang and Igor Babuschkin and Suchir Balaji and Shantanu Jain and William Saunders and Christopher Hesse and Andrew N. Carr and Jan Leike and Josh Achiam and Vedant Misra and Evan Morikawa and Alec Radford and Matthew Knight and Miles Brundage and Mira Murati and Katie Mayer and Peter Welinder and Bob McGrew and Dario Amodei and Sam McCandlish and Ilya Sutskever and Wojciech Zaremba},
  year={2021},
  eprint={2107.03374},
  archivePrefix={arXiv},
  primaryClass={cs.LG}
}

@article{hendrycks2021measuring,
  title={Measuring mathematical problem solving with the math dataset},
  author={Hendrycks, Dan and Burns, Collin and Kadavath, Saurav and Arora, Akul and Basart, Steven and Tang, Eric and Song, Dawn and Steinhardt, Jacob},
  journal={arXiv preprint arXiv:2103.03874},
  year={2021}
}

@article{yang2025qwen3,
  title={Qwen3 technical report},
  author={Yang, An and Li, Anfeng and Yang, Baosong and Zhang, Beichen and Hui, Binyuan and Zheng, Bo and Yu, Bowen and Gao, Chang and Huang, Chengen and Lv, Chenxu and others},
  journal={arXiv preprint arXiv:2505.09388},
  year={2025}
}

@article{grattafiori2024llama,
  title={The llama 3 herd of models},
  author={Grattafiori, Aaron and Dubey, Abhimanyu and Jauhri, Abhinav and Pandey, Abhinav and Kadian, Abhishek and Al-Dahle, Ahmad and Letman, Aiesha and Mathur, Akhil and Schelten, Alan and Vaughan, Alex and others},
  journal={arXiv preprint arXiv:2407.21783},
  year={2024}
}

@article{li2025eagle,
  title={Eagle-3: Scaling up inference acceleration of large language models via training-time test},
  author={Li, Yuhui and Wei, Fangyun and Zhang, Chao and Zhang, Hongyang},
  journal={arXiv preprint arXiv:2503.01840},
  year={2025}
}

@article{kojima2022large,
  title={Large language models are zero-shot reasoners},
  author={Kojima, Takeshi and Gu, Shixiang Shane and Reid, Machel and Matsuo, Yutaka and Iwasawa, Yusuke},
  journal={Advances in neural information processing systems},
  volume={35},
  pages={22199--22213},
  year={2022}
}

@article{snell2024scaling,
  title={Scaling llm test-time compute optimally can be more effective than scaling model parameters},
  author={Snell, Charlie and Lee, Jaehoon and Xu, Kelvin and Kumar, Aviral},
  journal={arXiv preprint arXiv:2408.03314},
  year={2024}
}

@article{welleck2024decoding,
  title={From decoding to meta-generation: Inference-time algorithms for large language models},
  author={Welleck, Sean and Bertsch, Amanda and Finlayson, Matthew and Schoelkopf, Hailey and Xie, Alex and Neubig, Graham and Kulikov, Ilia and Harchaoui, Zaid},
  journal={arXiv preprint arXiv:2406.16838},
  year={2024}
}

@misc{anthropic_claude_sonnet,
  title        = {Introducing Claude Sonnet 4.5},
  author       = {{Anthropic}},
  year         = {2025},
  howpublished = {\url{https://www.anthropic.com/news/claude-sonnet-4-5}},
}

@article{chen2023accelerating,
  title={Accelerating large language model decoding with speculative sampling},
  author={Chen, Charlie and Borgeaud, Sebastian and Irving, Geoffrey and Lespiau, Jean-Baptiste and Sifre, Laurent and Jumper, John},
  journal={arXiv preprint arXiv:2302.01318},
  year={2023}
}

@article{olsson2022context,
  title={In-context learning and induction heads},
  author={Olsson, Catherine and Elhage, Nelson and Nanda, Neel and Joseph, Nicholas and DasSarma, Nova and Henighan, Tom and Mann, Ben and Askell, Amanda and Bai, Yuntao and Chen, Anna and others},
  journal={arXiv preprint arXiv:2209.11895},
  year={2022}
}

@article{elhage2021mathematical,
  title={A mathematical framework for transformer circuits},
  author={Elhage, Nelson and Nanda, Neel and Olsson, Catherine and Henighan, Tom and Joseph, Nicholas and Mann, Ben and Askell, Amanda and Bai, Yuntao and Chen, Anna and Conerly, Tom and others},
  journal={Transformer Circuits Thread},
  volume={1},
  number={1},
  pages={12},
  year={2021}
}

@inproceedings{hendel2023context,
  title={In-Context Learning Creates Task Vectors},
  author={Hendel, Roee and Geva, Mor and Globerson, Amir},
  booktitle={Conference on Empirical Methods in Natural Language Processing},
year = {2023}
}

@article{eisenstadt2025overclocking,
  title={Overclocking LLM Reasoning: Monitoring and Controlling Thinking Path Lengths in LLMs},
  author={Eisenstadt, Roy and Zimerman, Itamar and Wolf, Lior},
  journal={arXiv preprint arXiv:2506.07240},
  year={2025}
}

@inproceedings{park2024linear,
  title={The Linear Representation Hypothesis and the Geometry of Large Language Models},
  author={Park, Kiho and Choe, Yo Joong and Veitch, Victor},
  booktitle={International Conference on Machine Learning},
  pages={39643--39666},
  year={2024},
  organization={PMLR}
}

@article{vaswani2017attention,
  title={Attention is all you need},
  author={Vaswani, Ashish and Shazeer, Noam and Parmar, Niki and Uszkoreit, Jakob and Jones, Llion and Gomez, Aidan N and Kaiser, {\L}ukasz and Polosukhin, Illia},
  journal={Advances in neural information processing systems},
  volume={30},
  year={2017}
}

@misc{zhou2024distillspecimprovingspeculativedecoding,
      title={DistillSpec: Improving Speculative Decoding via Knowledge Distillation}, 
      author={Yongchao Zhou and Kaifeng Lyu and Ankit Singh Rawat and Aditya Krishna Menon and Afshin Rostamizadeh and Sanjiv Kumar and Jean-François Kagy and Rishabh Agarwal},
      year={2024},
      eprint={2310.08461},
      archivePrefix={arXiv},
      primaryClass={cs.CL},
      url={https://arxiv.org/abs/2310.08461}, 
}

@inproceedings{quan-etal-2025-rasd,
    title = "{RASD}: Retrieval-Augmented Speculative Decoding",
    author = "Quan, Guofeng  and
      Feng, Wenfeng  and
      Hao, Chuzhan  and
      Jiang, Guochao  and
      Zhang, Yuewei  and
      Wang, Hao Henry",
    editor = "Che, Wanxiang  and
      Nabende, Joyce  and
      Shutova, Ekaterina  and
      Pilehvar, Mohammad Taher",
    booktitle = "Findings of the Association for Computational Linguistics: ACL 2025",
    month = jul,
    year = "2025",
    address = "Vienna, Austria",
    publisher = "Association for Computational Linguistics",
    url = "https://aclanthology.org/2025.findings-acl.320/",
    doi = "10.18653/v1/2025.findings-acl.320",
    pages = "6167--6177",
    ISBN = "979-8-89176-256-5"
}

@inproceedings{hu2025sam,
  title={Sam decoding: Speculative decoding via suffix automaton},
  author={Hu, Yuxuan and Wang, Ke and Zhang, Xiaokang and Zhang, Fanjin and Li, Cuiping and Chen, Hong and Zhang, Jing},
  booktitle={Proceedings of the 63rd Annual Meeting of the Association for Computational Linguistics (Volume 1: Long Papers)},
  pages={12187--12204},
  year={2025}
}

@inproceedings{zhao2024lookahead,
  title={Lookahead: An inference acceleration framework for large language model with lossless generation accuracy},
  author={Zhao, Yao and Xie, Zhitian and Liang, Chen and Zhuang, Chenyi and Gu, Jinjie},
  booktitle={Proceedings of the 30th ACM SIGKDD Conference on Knowledge Discovery and Data Mining},
  pages={6344--6355},
  year={2024}
}

@article{Hou2025BanditSpecAS,
  title={BanditSpec: Adaptive Speculative Decoding via Bandit Algorithms},
  author={Yunlong Hou and Fengzhuo Zhang and Cunxiao Du and Xuan Zhang and Jiachun Pan and Tianyu Pang and Chao Du and Vincent Y. F. Tan and Zhuoran Yang},
  journal={ArXiv},
  year={2025},
  volume={abs/2505.15141},
  url={https://api.semanticscholar.org/CorpusID:278782841}
}

@inproceedings{See2017GetTT,
  title={Get To The Point: Summarization with Pointer-Generator Networks},
  author={A. See and Peter J. Liu and Christopher D. Manning},
  booktitle={Annual Meeting of the Association for Computational Linguistics},
  year={2017},
  url={https://api.semanticscholar.org/CorpusID:8314118}
}

@article{Hermann2015TeachingMT,
  title={Teaching Machines to Read and Comprehend},
  author={Karl Moritz Hermann and Tom{\'a}s Kocisk{\'y} and Edward Grefenstette and Lasse Espeholt and Will Kay and Mustafa Suleyman and Phil Blunsom},
  journal={ArXiv},
  year={2015},
  volume={abs/1506.03340},
  url={https://api.semanticscholar.org/CorpusID:6203757}
}


\appendix


\section{Impact of Sampling Temperature\label{app:impact_of_temp}}

All main experiments in the paper are conducted with temperature $T=0.0$, following the standard setup commonly used in SD evaluations~\cite{leviathan2023fast,chen2023accelerating}. Since SD performance depends on the agreement between the draft and target model distributions, increasing the sampling temperature generally reduces acceptance rates and therefore lowers the achievable speedup.

To evaluate the robustness of \methodname under different decoding configurations, we additionally measure throughput across a range of sampling temperatures on HumanEval using LLaMA-3.1-8B-Instruct. Results are reported in Table~\ref{tab:temperature_results}.

\begin{table}[h]
\centering
\small
\begin{tabular}{lcccc}
\toprule
Temperature & Base Tok/s & EAGLE3 Tok/s & SwitchSD Tok/s & Speedup (SwitchSD) \\
\midrule
0.0 & 38.3 & 87.5 & \textbf{98.6} & \textbf{2.57$\times$} \\
0.2 & 36.3 & 83.1 & \textbf{88.5} & \textbf{2.44$\times$} \\
0.4 & 36.7 & 80.6 & \textbf{83.9} & \textbf{2.29$\times$} \\
0.6 & 36.2 & 79.7 & \textbf{80.6} & \textbf{2.22$\times$} \\
0.8 & 36.4 & 73.3 & \textbf{74.9} & \textbf{2.05$\times$} \\
1.0 & 36.8 & 53.9 & \textbf{56.5} & \textbf{1.54$\times$} \\
\bottomrule\\
\end{tabular}
\caption{Throughput across different sampling temperatures on HumanEval using LLaMA-3.1-8B.}
\label{tab:temperature_results}
\end{table}

Overall, \methodname consistently outperforms EAGLE3 across all evaluated temperatures. As expected, the performance gap gradually decreases as temperature increases. Higher temperatures induce more stochastic and diverse generations, reducing repetition frequency and lowering speculative token acceptance rates for both copy-based and neural drafting strategies. Nevertheless, even at $T=1.0$, \methodname still achieves a substantial $1.54\times$ speedup over vanilla decoding while maintaining an advantage over EAGLE3.

\section{Additional Comparison to Context-Exploitation Baselines}
\label{app:additional_baselines}

To further evaluate the effectiveness of \methodname against non-parametric context-exploitation methods, we compare out method against several additional baselines that rely on context copying and n-gram retrieval mechanisms, including Prompt Lookup Decoding (PLD) \citep{saxena2023prompt}, Sam decoding (SAMD) \citep{hu2025sam}, and Lookahead Decoding \citep{zhao2024lookahead}. All experiments are conducted using LLaMA-3.1-8B-Instruct.

\begin{table}[h]
\centering
\small
\begin{tabular}{llccc}
\toprule
Dataset & Method & Tok/s & Speedup & MAL (copy/EAGLE) \\
\midrule
\multirow{4}{*}{HumanEval}
& PLD & 53.0 & 1.38$\times$ & 1.59 \\
& SAMD & 48.4 & 1.26$\times$ & 1.63 \\
& Lookahead & 77.7 & 2.03$\times$ & 3.17 \\
& \methodname & \textbf{98.63} & \textbf{2.58$\times$} & \textbf{5.88 / 2.93} \\
\midrule
\multirow{4}{*}{Math500}
& PLD & 53.1 & 1.37$\times$ & 1.54 \\
& SAMD & 52.3 & 1.35$\times$ & 1.60 \\
& Lookahead & 80.3 & 2.07$\times$ & 3.10 \\
& \methodname & \textbf{88.74} & \textbf{2.29$\times$} & \textbf{5.39 / 2.41} \\
\midrule
\multirow{4}{*}{CNN/DailyMail}
& PLD & 50.7 & 1.33$\times$ & 1.51 \\
& SAMD & 44.7 & 1.17$\times$ & 1.53 \\
& Lookahead & 63.9 & 1.67$\times$ & 2.85 \\
& \methodname & \textbf{77.00} & \textbf{2.01$\times$} & \textbf{2.84 / 2.46} \\
\bottomrule \\
\end{tabular}
\caption{Comparison against additional context-exploitation SD baselines using LLaMA-3.1-8B-Instruct. MAL denotes mean accepted length.}
\label{tab:additional_context_baselines}
\end{table}

Results are presented in Table~\ref{tab:additional_context_baselines}. Across all datasets, \methodname consistently achieves the highest throughput and speedup. In particular, while PLD and SAMD often produce short accepted copy spans, and Lookahead improves acceptance through more aggressive retrieval strategies, \methodname obtains substantially longer accepted copy continuations while preserving the benefits of neural drafting through EAGLE3. These results further demonstrate that selectively activating copying based on intrinsic copy-intent signals is substantially more effective than indiscriminate context exploitation.

\section{CopyDiversity Dataset \label{app:CopyDataset}}

CopyDiversity is a curated dataset designed to capture a wide spectrum of copy behavior in language model generation, ranging from strict verbatim reproduction to fully open-ended responses. The dataset contains 800 training samples spanning 17 prompt types, intentionally constructed to vary the degree to which copying from the input context is expected or beneficial. This diversity enables systematic analysis of copy intent and accidental repetition under realistic generation conditions. The dataset consists of 800 train examples and 200 validation examples.

\textbf{Heavy Copy} prompts (10.6\% of the data) explicitly require verbatim reproduction of provided content. Representative examples include:
\begin{itemize}

    \item \emph{JSON Echo}: \textit{User profile:}
\begin{verbatim}
{
  "name": "Bob",
  "age": 45,
  "city": "London",
  "skills": ["SQL", "JavaScript"]
}
\end{verbatim}
\textit{Output the same JSON with proper formatting.}
\textit{Copy the JSON exactly.}
    \item \emph{Quote Extraction}: \textit{"Text: The Nile River is the longest river in the world. It flows through 11 countries in northeastern Africa. Gravity is the force that attracts objects toward each other. On Earth, it gives weight to physical objects. Photosynthesis converts sunlight into chemical energy. Plants use this process to produce glucose and oxygen. Identify and quote exactly 3 important sentences from the passage:"}
\end{itemize}
In these settings, faithful copying is essential for correctness.

\textbf{Medium Copy} prompts (11.7\% of the data) require partial reuse of the input content, typically combined with light transformation or selective extraction. Example prompts include:
\begin{itemize}
    \item \emph{Definition Recall}:
\begin{verbatim}
Key terms:
Graph: A collection of nodes (vertices) connected by edges representing 
relationships between elements.
Algorithm: A step-by-step procedure for solving a problem or accomplishing
a task. Interpreter: A program that executes source code line by line without 
prior compilation.
Linked List: A linear data structure where elements are stored in nodes 
connected by pointers.
Server: A computer program or device that provides services and resources to
other programs or devices.

Define each term using the exact definitions above:
1. What is a graph?
2. What is an algorithm?
3. What is an interpreter?
4. What is a linked list?
5. What is a server?
\end{verbatim}
    \item \emph{Summarize and Quote}: \textit{"Article: The Amazon rainforest produces 20\% of the world's oxygen. It is home to millions of species. The Pacific Ocean is the largest and deepest ocean on Earth. It covers more than 60 million square miles. Summarize briefly, then include 2 exact quotes from the passage."}
\end{itemize}
These tasks exhibit mixed behavior, where copying is necessary but must be integrated with generation.

\textbf{Light or No Copy} prompts (77.7\% of the data) are open-ended and do not require verbatim reuse of the input. Examples include:
\begin{itemize}
    \item \emph{Creative Writing}: \textit{Describe a world where dreams can be shared between people.}
    \item \emph{Reasoning Problem}: \textit{A lily pad doubles in size every day. If it takes 48 days to cover a lake, on which day is the lake half covered?}
\end{itemize}
In these cases, copying is generally incidental and often undesirable.

By explicitly balancing prompt types across these three regimes, CopyDiversity supports fine-grained evaluation of copy intent detection and copy-aware decoding strategies across diverse and realistic use cases.

\section{Implementation Details}
We use PyTorch for all experiments. Across all models, \methodname generates up to 10 copy-based candidate sequences per step, each with a maximum length of 20 tokens, which are verified in parallel by the target model. All experiments were conducted on NVIDIA H100 GPUs, with individual runs ranging from 45 minutes to 17 hours, depending on model size and dataset. 

\section{Discussion: What Does the Probe Learn?\label{sec:discussion}}
Learning copy intent is inherently challenging, since there is no direct ground-truth annotation for this behavior. We therefore train the probe with weak supervision: automatically identified long repetitions are used as proxy labels for copy-oriented behavior. A natural concern is that the resulting probe may learn surface-level repetition rather than the copy-intent signal it is designed to capture. Our design addresses this concern in three ways:

\textbf{(i) Weak supervision.} Repeated spans are used only as a supervisory proxy, not as the definition of copy intent. Long verbatim continuations provide high-precision examples of situations in which copying is likely to be useful. 

\textbf{(ii) Representational Bottleneck.}
While verbatim spans provide the supervisory labels, the probe's input is strictly constrained to the hidden representation of a single token at a single timestep. This creates a significant information bottleneck: the probe does not have access to the speculative buffer or the context window as a sequence. Consequently, the probe cannot be performing a post-hoc string match. To achieve high precision, it must isolate a specialized "copy-intent" signature distilled within the latent space, a signal that indicates the model has transitioned its internal logic from novel synthesis to contextual reproduction.

\textbf{(iii) Data quality.} The training data is constructed from diverse, model-generated completions with varying degrees of copying, rather than from arbitrary repeated tokens. This exposes the probe to realistic decoding contexts and encourages generalization beyond simple string matching.

\textbf{(iv) Regularized representation.} The probe is a lightweight classifier trained on frozen LLM hidden representations, rather than a module that directly observes the full context, token embeddings, or KV cache.

Finally, the empirical results support this interpretation. Compared with heuristic copy methods such as CopySpec and BanditSpec, \methodname invokes copying less frequently but achieves longer accepted copy spans and higher throughput. Moreover, when conditioning on positions where a copy candidate exists, probe-active states yield substantially longer accepted spans, whereas probe-inactive states mostly lead to short or zero-token acceptances. Thus, the probe is best understood as learning copy-readiness: an internal signal that predicts when a surface repetition is likely to match the target model's continuation.

\end{document}